\documentclass[nonblindrev]{informs3} 

\usepackage{balance}
\usepackage{amsmath}
\usepackage[tight,footnotesize]{subfigure}
\usepackage{graphicx}
\usepackage{bm}
\usepackage{algorithm}
\usepackage[noend]{algorithmic}
\usepackage{multirow}
\usepackage{slashbox}
\usepackage{caption}
\usepackage{amsfonts}

\OneAndAHalfSpacedXII 

\usepackage{natbib}
 \bibpunct[, ]{(}{)}{,}{a}{}{,}%
 \def\bibfont{\small}%
\TheoremsNumberedThrough     

\EquationsNumberedThrough    

\MANUSCRIPTNO{2021} 

\begin{document}


\RUNAUTHOR{Tang et al.}

\RUNTITLE{Adaptive Cascade Submodular Maximization}

\TITLE{Adaptive Cascade Submodular Maximization}

\ARTICLEAUTHORS{%
\AUTHOR{Shaojie Tang}
\AFF{Naveen Jindal School of Management, The University of Texas at Dallas}
\AUTHOR{Jing Yuan}
\AFF{Department of Computer Science, The University of Texas at Dallas}
} 

\ABSTRACT{In this paper, we propose and study the cascade submodular maximization problem under the adaptive setting. The input of our problem is a set of items, each item is in a particular state (i.e., the marginal contribution of an item) which is  drawn from a known probability distribution. However, we can not know its actual state before selecting it. As compared with existing studies on stochastic submodular maximization, one unique setting of our problem is that each item is associated with a continuation probability which represents the probability that one is allowed to  continue to select the next item after selecting the current one. Intuitively, this term captures the externality  of selecting one item to all its subsequent items in terms of the opportunity of being selected.
Therefore, the actual set of items that can be selected by a policy depends on the specific ordering it adopts to select items, this makes our problem fundamentally different from classical submodular set optimization problems. Our objective is to identify the best sequence of selecting items so as to maximize the expected utility of the selected items. We propose a class of stochastic utility functions, \emph{adaptive cascade submodular functions}, and show that the objective functions in many practical application domains satisfy adaptive cascade submodularity.
Then we develop a $0.12$ approximation algorithm to the adaptive cascade submodular maximization problem.
}


\maketitle

%
\section{Introduction}
\label{sec:intro}
Submodular maximization has been extensively studied in the literature \citep{nemhauser1978analysis}. Their objective is to select a group of items that maximize a submodular utility function subject to various constraints. Recently, \citet{golovin2011adaptive} propose the problem of  adaptive submodular maximization, a natural stochastic variant of the classical submodular maximization. In particular,  they assume that each item is associated with a particular state which is drawn from a known distribution, the only way to know an item's state is to select that item. As compared with the classical submodular maximization, feasible solutions are now policies instead of subsets: the action taken in each step depends on the observations from the previous steps. For example, a typical adaptive policy works as follows: in each step, we select an item, get to see its actual state, then adaptively select the next item based on these observations and so on. They show that an  adaptive greedy policy achieves a $1-1/e$ approximation ratio when maximizing an adaptive submodular and adaptive monotone utility function subject to a cardinality constraint. Our problem setting is similar to theirs in that we are also interested in selecting a sequence of items adaptively so as to maximize the expected utility. However, one unique setting of our model is that each item is assigned a \emph{continuation probability} \citep{craswell2008experimental,tangyuan2020}. The continuation probability of an item, say $i$, is defined as the probability that one is allowed to select the next item  after $i$ is being selected. The probabilistic continuation setting allows us to capture the scenario where the selecting process could be terminated prematurely. It can be seen that selecting an item with low continuation probability decreases the chance of its subsequent items being selected.     Therefore, the actual set of items that can be selected by a policy depends on the specific ordering it adopts to select items. Our setting is motivated by many real-world applications in machine learning, economics, and operations management. We next give three real-world examples that fit our setting.

\textbf{Example 1:}  Taking sponsored search advertising as one example,  one challenge faced by ad-networks is to select a sequence of advertisements to display to an online user. It is often assumed that the visibility of an advertisement is negatively impacted by the appearances of its preceding advertisements, e.g., the user is less likely to view a new ad after viewing too many other advertisements. One common way to capture this effect is to introduce the continuation probability for each advertisement. In particular, \citet{tang2020optimizing} assume that the users scan through the ads in order, after viewing one advertisement, users decide probabilistically whether to click it, as well as whether to continue the scanning process with the ad specific continuation probability. As a result, the user could terminate the ad session prematurely in a probabilistic manner. Assume the revenue generated from one click is known, the goal of ad-network is to adaptively select a sequence of ads to maximize the expected revenue.

\textbf{Example 2:}  Our second practical application is sequential product recommendation \citep{flores2019assortment}. One important task of online retailers such as Amazon is to recommend a list of products to each online user. Given a list of recommended products, users scan through them in order, after browsing one product, users decide probabilistically whether to browse the next product with a product-specific continuation probability. After browsing some products, users make a purchase decision within these products (including no purchase option). Similar to the first application, the recommendation process stops immediately whenever the user decides not to browse the next product. Assume each product is associated with a revenue, the goal of the online retailer is to adaptively recommend a group of products to maximize the expected revenue.

\textbf{Example 3:} The third example is pool-based active learning \citep{golovin2011adaptive}. For example, given a set of possible hypotheses about
the profile of an online user, we let the user conduct an online survey to rule out inconsistent hypotheses. One popular way to build the online survey is called paging design \citep{tangyuan2020} where only one question is displayed to the user per page. After finishing a question, the user can click the ``Next''
button to view the next question or click ``Exit'' to terminate the survey process probabilistically.  Our goal is to adaptively choose a sequence of survey questions to infer the user's profile as accurate as possible. In Section \ref{sec:experiment}, we use pool-based active learning as an example application to evaluate the performance of our algorithms.

One crucial point in the above applications is that the value of a group of items depends not only on the items belong to that group, but also on the specific ordering of those items.  This makes our problem different from set optimization problems as we seek  a best sequence of items  while considering the externality of one item to its subsequent items in terms of the chance of being selected. Although sequence selection has attracted increasing attention these days, most of existing results do not apply to our problem. As will be discussed in Section \ref{sec:related}, our utility function does not satisfy the property of ``non-decreasing'', a common assumption made in many existing studies. Moreover, while the majority of prior research considers non-adaptive setting, we focus on the adaptive setting, where we are allowed to dynamically adjust the selecting strategy based on the current observations. To make our problem approachable, we restrict our attention to a class of stochastic utility functions, \emph{adaptive cascade submodular functions}. Intuitively, any adaptive cascade submodular function  must satisfy \emph{diminishing return} condition under the adaptive setting.  We show that the objective functions in several practical application domains satisfy adaptive cascade submodularity.   We propose a simple algorithm that achieves a  $0.12$ approximation ratio. We conduct extensive experiments to evaluate the performance of our solutions in the context of pool-based active learning. The experiment results validate the theoretical analysis of our algorithm.

\section{Related Work}
\label{sec:related}
Submodular maximization has been extensively studied in the literature \citep{nemhauser1978analysis}. However, most of existing studies focus on set optimization problems whose objective is to select set of items that maximizes a submodular utility function. Our focus is on identifying the best sequence of items so as to maximize the expected utility. Although our paper focuses on the adaptive setting, we first review some important studies in the filed of non-adaptive sequence selection. Recently, \citet{streeter2009online} considered the sequence optimization problem prompted by applications such as online resource allocation. They defined the properties of monotonicity and submodularity over sequences instead of sets. \citet{alaei2010maximizing} introduced the term of sequence submodularity and sequence-non-decreasing. \citet{tschiatschek2017selecting} and \citet{mitrovic2018submodularity} defined the utility of a sequence over the edges of a directed graph connecting the items together with a submodular function. However, their results do not apply to our setting since our utility function does not satisfy the property of ``sequence monotonicity''. Intuitively, under our setting, adding an item to an existing sequence could decrease the utility of the original sequence. For example, assume there is a sequence $S$ with positive utility, as well as an item $i$ with zero utility and zero continuation probability. Consider a new sequence that concatenates $i$ and $S$ ($i$ is placed ahead of $S$), it is easy to verify that the utility of the new sequence is zero which is smaller than the utility of $S$. \citet{zhang2015string} propose the concept of string submodularity. They provide a set of data-dependent approximation bounds for a greedy strategy.  It turns out that our utility function  under the non-adaptive setting is string submodular, however, the worst-case performance of the greedy strategy \citep{zhang2015string} is arbitrarily bad in our setting. \citet{tangyuan2020} propose and study the cascade submodular maximization problem under the non-adaptive setting. They propose a series of approximation algorithms in the context of quiz design. Note that all studies previously mentioned restrict themselves to the non-adaptive setting where a sequence must be selected all at once. Only recently, \citet{mitrovic2019adaptive} extend the previous studies to the adaptive setting and propose the concept of adaptive sequence submodularity. They follow \citep{tschiatschek2017selecting} to build their  basic model. Our study is different from theirs in that our utility function is defined over subsequences, instead of graphs \citep{mitrovic2019adaptive}, making their results not applicable to our setting.

Our work is also closely related to stochastic submodular maximization \citep{asadpour2016maximizing,golovin2011adaptive}. \citet{golovin2011adaptive} extend submodularity to adaptive policies and propose the concept of  adaptive submodularity. They show that the greedy adaptive strategy achieves a $1-1/e$ approximation ratio for adaptive submodular maximization subject to a cardinality constraint. \citet{tang2020beyond} develops the first approximation algorithm for maximizing a non-monotone adaptive submodular function.
 In this work, we generalize the concept of adaptive submodularity to functions over sequences instead of sets, and introduce the concept of adaptive cascade submodular functions.  As mentioned earlier,  our model allows us to capture the scenario where the selecting
process could be terminated prematurely. We develop an adaptive policy that achieves a $0.12$ approximation ratio for solving sequence selection problems with an adaptive cascade submodular and monotone function.

\section{Preliminaries}
We first introduce some notations and define the general class of \emph{adaptive cascade submodular}  functions. In the rest of this paper, let $[n]$ denote the set $\{1, 2, \cdots, n\}$, and we use $|S|$ to denote the cardinality of a set or a sequence $S$.

\subsection{Items and States} Let $E$ denote the entire set of $m$ items, and each item $i\in E$ is in a particular state that belongs to  a set $O$ of possible states.  Denote by $\phi: E\rightarrow O$  a realization of the states of items.   Let $\Phi=\{\Phi_i\mid i\in E\}$ be a random realization where $\Phi_i \in O$ denotes a random realization of $i$.   After selecting $i$, its actual state $\Phi_i$ is discovered. Let $\mathcal{U}$ denote the set of all realizations, we assume there is a known prior probability distribution $p(\phi)=\{\Pr[\Phi=\phi]: \phi\in \mathcal{U}\}$ over realizations. In addition, there is a vector $\delta=\{\delta_i \mid i\in E\}$ where $\delta_i$ denotes the  \emph{continuation probability} of item $i\in E$, i.e., it represents the  probability that one is allowed to  continue to select the next item after selecting $i$. We are interested in selecting a group of items adaptively as follows: we start by selecting the first item, say $i\in E$, observe its state $\Phi_i$, then with probability $\delta_i$, we continue to select the next item  and observe its state, otherwise we terminate the selecting process, and so on. During the selecting process, we say the current process is \emph{live} if we can continue to select the next item, otherwise we say this process is \emph{dead}. Thus, the probability of a selecting process to be live after $i$ is being selected  is $\delta_i$.  After each selection, we denote by a \emph{partial realization} $\psi$ the observations made so far: $\psi$ is a function from some subset (i.e., those items which are selected so far) of $E$ to their states. We define the \emph{domain} of $\psi$ as the subset of items involved in $\psi$. Given a realization $\phi$ and a partial realization $\psi$, we say $\psi$ is consistent with $\phi$ if they are equal everywhere in the domain of $\psi$. We write $\phi \sim \psi$ in this case. We say that $\psi$  is a \emph{subrealization} of  $\psi'$ if $\mathrm{dom}(\psi) \subseteq \mathrm{dom}(\psi')$ and they are equal everywhere in $\mathrm{dom}(\psi)$.  In this case we write $\psi \subseteq \psi'$. We use $p(\phi\mid \psi)$ to denote the conditional distribution over realizations given a partial realization $\psi$: $p(\phi\mid \psi) =\Pr[\Phi=\phi\mid \Phi \sim\psi ]$.

\subsection{Policies and Problem Formulation} Any adaptive strategy of selecting items can be represented using a function $\pi$  from a set of partial
realizations to $E$, specifying  which item to select next, if the current selecting process is still live, given the current observations. Given any policy $\pi$ and realization $\phi$, we say $\pi$ \emph{adopts} $S_{\pi, \phi}$ under realization $\phi$ if $S_{\pi, \phi}$ is the longest possible sequence of items that can be selected by $\pi$ under realization $\phi$. Intuitively, by following $\pi$, one can successfully select all items in  $S_{\pi, \phi}$ under $\phi$ if the selecting process is never dead. By abuse of notation,  we use the same notation $S_{\pi, \phi}$  to denote the set of items in $S_{\pi, \phi}$.   Given a sequence $S_{\pi, \phi}$, let $S^{(k)}_{\pi, \phi}$ denote the prefix of $S_{\pi, \phi}$ of length $k$  and let $S_{\pi, \phi}^{k}$ denote the $k$-th item in $S_{\pi, \phi}$ for any $k\in[m]$.  It follows that  under realization $\phi$, $\pi$ selects all and only items from  $S^{(k)}_{\pi, \phi}$  with probability $(1-\delta_k)\prod_{i\in S^{(k-1)}_{\pi, \phi}}\delta_i $.

For notational convenience, define $S^{(0)}_{\pi, \Phi}=\emptyset$ for any $\pi$ and $\Phi$. We next introduce a utility function $f$ from a subset of items and their states to a non-negative real number: $f: 2^{E}\times O^E\rightarrow \mathbb{R}_{\geq0}$. The expected utility of a policy $\pi$ under realization $\phi$ is
\[\sum_{k\in [|S_{\pi, \phi}|]} (1-\delta_{S_{\pi, \phi}^{k}})\prod_{i\in S_{\pi, \phi}^{(k-1)}}\delta_i f(S^{(k)}_{\pi, \phi}, \phi)\]
 Based on this notation, we define the expected  utility $f_{avg}(\pi)$ of a policy $\pi$ as
\begin{equation}
f_{avg}(\pi)=\mathbb{E}_{\Phi\sim p(\phi)}[\sum_{k\in [|S_{\pi, \Phi}|]} (1-\delta_{S_{\pi, \Phi}^{k}})\prod_{i\in S^{(k-1)}_{\pi, \Phi}}\delta_i f(S^{(k)}_{\pi, \Phi}, \Phi)]
\end{equation}

Our goal is to find a policy  $\pi^{opt}$ that maximizes the expected utility:
\[\pi^{opt} \in \arg\max_{\pi} f_{avg}(\pi)\]

\subsection{Adaptive Cascade  Submodularity and Monotonicity}
We first review two concepts which are defined over  set functions. For notational convenience, let $h(\psi)=\mathbb{E}_{\Phi\sim p(\phi\mid \psi)}[f(\mathrm{dom}(\psi), \Phi)]$ denote the utility of $\mathrm{dom}(\psi)$ under partial realization $\psi$.
\begin{definition} \citep{golovin2011adaptive}[Adaptive Submodularity]
\label{def:3}
A set function $f$ is  adaptive  submodular with respect to a prior distribution $ p(\phi)$, if for any  two partial realizations $\psi$ and $\psi'$ such that $\psi \subseteq \psi'$, and any item $i\in E\setminus \mathrm{dom}(\psi')$, the following holds:
\begin{eqnarray}
&& \mathbb{E}_{\Phi\sim p(\phi\mid \psi)}[f(\{i\} \cup \mathrm{dom}(\psi), \Phi)]-h(\psi) \nonumber\\
&&\quad\quad\quad \geq  \mathbb{E}_{\Phi\sim p(\phi\mid \psi')}[f(\{i\} \cup \mathrm{dom}(\psi'), \Phi)]- h(\psi')
\end{eqnarray}
\end{definition}

\begin{definition} \citep{golovin2011adaptive}[Adaptive Monotonicity]
\label{def:2}
A set function $f$ is  adaptive  monotone with respect to a prior distribution $ p(\phi)$, if for any  partial realization $\psi$, and any item $i\in E\setminus \mathrm{dom}(\psi)$, the following holds:
\begin{equation}
 \mathbb{E}_{\Phi\sim p(\phi\mid \psi)}[f(\{i\} \cup \mathrm{dom}(\psi), \Phi)]-h(\psi) \geq 0
\end{equation}
\end{definition}

We next introduce the notation of \emph{adaptive cascade submodularity}. For any subset of items $V\subseteq E$, let  $\Omega(V)$  denote the set of  policies which are allowed to select items only from $V$.  It clear that $\Omega(V)  \subseteq \Omega(V')$ for any $V \subseteq V'$.
\begin{definition}[Adaptive Cascade Submodularity]
\label{def:1}
A function $ f$ is  adaptive cascade submodular with respect to a prior distribution $ p(\phi)$, if for any two partial realizations $\psi$ and $\psi'$ such that $\psi\subseteq \psi'$, and any subset of items $V\subseteq E\setminus \mathrm{dom}(\psi')$, the following holds for any $\delta$:
\begin{eqnarray}
&&\max_{\pi\in\Omega(V)} f_{avg}(\pi\cup \mathrm{dom}(\psi)\mid \psi)-h(\psi)\nonumber \\
&&\quad\quad\quad\geq \max_{\pi\in\Omega(V)} f_{avg}(\pi\cup \mathrm{dom}(\psi')\mid \psi')-h(\psi')\nonumber
\end{eqnarray}
where $f_{avg}(\pi\cup \mathrm{dom}(\psi)\mid \psi)=$
{\small
\begin{equation}
\mathbb{E}_{\Phi\sim p(\phi\mid \psi)}[\sum_{k\in [|S_{\pi, \Phi}|]} (1-\delta_{S_{\pi, \Phi}^{k}})\prod_{i\in S_{\pi, \Phi}^{(k-1)}}\delta_i f(S_{\pi, \Phi}^{(k)}\cup \mathrm{dom}(\psi), \Phi)] \label{eq:7777}
\end{equation} }denote the conditional expected utility of a policy that first selects $\mathrm{dom}(\psi)$, then runs $\pi$, conditioned on a partial realization $\psi$.
\end{definition}

We next show that  
if $ f$ is adaptive monotone and adaptive cascade submodular with respect to a prior distribution $ p(\phi)$, then  $f$ is adaptive submodular with respect to the same distribution.  Although the other direction is not necessarily  true, that is,   adaptive submodularity does not imply adaptive cascade submodularity, we find that many well studied adaptive submodular functions are also adaptive cascade submodular. In fact, it is easy to show that if the variables $\{\Phi_i\mid i\in E\}$ are independent, then $f$ is adaptive cascade submodular. One such example is sensor placement \citep{krause2007near} where deployed sensors are assumed to fail probabilistically and independently. In other applications including influence maximization and  pool-based active learning \citep{golovin2011adaptive} where $\{\Phi_i\mid i\in E\}$ are not independent, their utility functions also satisfy  adaptive cascade submodularity.

\begin{lemma}
\label{lem:4}
If  $f$ is adaptive monotone and adaptive cascade submodular  with respect to a prior distribution $ p(\phi)$, then $f$ is adaptive submodular with respect to  $ p(\phi)$.
\end{lemma}
\emph{Proof:}  Since $ f$ is  adaptive cascade submodular, we have
\begin{eqnarray}
&&\max_{\pi\in\Omega(V)} f_{avg}(\pi\cup \mathrm{dom}(\psi)\mid \psi)-h(\psi) \nonumber \\
&&\quad\quad \geq \max_{\pi\in\Omega(V)} f_{avg}(\pi\cup \mathrm{dom}(\psi')\mid \psi')-h(\psi') \label{Eq:999}
 \end{eqnarray}
 for any two partial realizations $\psi$ and $\psi'$ such that $\psi\subseteq \psi'$, and any subset of items $V\subseteq E$, according to Definition \ref{def:1}. Consider a singleton $V=\{i\}$ for any $i\in E\setminus \mathrm{dom}(\psi')$, the strategy set $\Omega(\{i\})$ contains only two strategies for any partial realization $\psi$: selecting $i$ or $\emptyset$. Due to $f$ is adaptive monotone, we have $\max_{\pi\in\Omega(\{i\})} f_{avg}(\pi\cup \mathrm{dom}(\psi)\mid \psi) = f_{avg}(\pi\cup \{i\}\mid \psi)$ for any $i\in E\setminus \mathrm{dom}(\psi')$ and partial realization $\psi$. Condition (\ref{Eq:999}) can be simplified as
 \begin{equation}
\label{Eq:666} f_{avg}(\pi\cup \{i\}\mid \psi)-h(\psi) \geq  f_{avg}(\pi\cup \{i\}\mid \psi')-h(\psi')
 \end{equation}
  for any two partial realizations $\psi$ and $\psi'$ such that $\psi\subseteq \psi'$, and any $i\in E\setminus \mathrm{dom}(\psi')$. According to (\ref{eq:7777}), we have $f_{avg}(\pi\cup \{i\}\mid \psi)=\mathbb{E}_{\Phi\sim  p(\phi\mid \psi)}[f(\{i\}\cup \mathrm{dom}(\psi), \Phi)]$ for any partial realization $\psi$. It follows that (\ref{Eq:666}) can be rewritten as
 \begin{eqnarray}
&& \mathbb{E}_{\Phi\sim  p(\phi\mid \psi)}[f(\{i\}\cup \mathrm{dom}(\psi), \Phi)]-h(\psi) \nonumber \\
&& \geq  \mathbb{E}_{\Phi\sim p(\phi\mid \psi')}[f(\{i\}\cup \mathrm{dom}(\psi'), \Phi)]-h(\psi') \label{Eq:6667}\end{eqnarray}   for any two partial realizations $\psi$ and $\psi'$ such that $\psi\subseteq \psi'$, and any $i\in E\setminus \mathrm{dom}(\psi')$. According to Definition \ref{def:3}, (\ref{Eq:6667}) implies that $f$ is adaptive submodular with respect to  $ p(\phi)$. $\Box$

\section{The Adaptive Greedy Plus Policy}
In this section, we propose an adaptive policy which achieves a constant approximation ratio for maximizing an adaptive cascade submodular function.
\subsection{Technical Lemmas}
Before presenting our algorithm, we first provide some additional notations and technical lemmas that will be used to design and analyze the proposed algorithm. We start by introducing the concept of \emph{reachability}.   
\begin{definition}[Reachability]
Given a sequence of items $S$ and any $k\in[|S|]$, we define the reachability  of its $k$-th item $S^{k}$ as  $\prod_{i\in S^{(k-1)}}\delta_i$, e.g., it represents the probability of $S^{k}$ being selected given that $S$ is adopted. For notational convenience, assume $\delta_{S^{0}}=1$.
\end{definition}

Based on the notion of reachability, we next introduce the concepts of \emph{$\rho$-reachable sequence}, \emph{strongly $\rho$-reachable sequence},   \emph{$\rho$-reachable policy}, and  \emph{strongly $\rho$-reachable policy}.

\begin{definition}[$\rho$-reachable Sequence]
 For any $\rho\in[0,1]$, we say a sequence $S$ is \emph{$\rho$-reachable}  if the reachability of all items in $S$ is at least $\rho$, or in equivalent, the reachability of the last item of $S$ is at least $\rho$, e.g.,  $\prod_{i\in S^{(|S|-1)}}\delta_i \geq \rho$.
\end{definition}

Based on the above definition, it can be seen that if we adopt a $\rho$-reachable sequence $S$, then the entire $S$ can be selected with probability at least $\rho$.

\begin{definition}[$\rho$-reachable  Policy]
 For any $\rho\in[0,1]$, we say a policy $\pi$ is  \emph{$\rho$-reachable} if  for any realization $\phi$, it holds that $\prod_{i\in S^{(|S_{\pi,\phi}|-1)}_{\pi,\phi}} \delta_i \geq \rho$. That is, $\pi$ only adopts  $\rho$-reachable sequence. Let $\Omega(\rho)$ denote the set of all $\rho$-reachable policies.
\end{definition}


\begin{definition}[Strongly $\rho$-reachable Sequence]
\label{def:2sequence}
 For any $\rho\in[0,1]$, we say a sequence $S$ is \emph{strongly $\rho$-reachable}  if $\prod_{i\in S} \delta_i \geq \rho$. \end{definition}

\begin{definition}[Strongly $\rho$-reachable Policy]
\label{def:2policy}
 For any $\rho\in[0,1]$, we say a policy $\pi$ is  \emph{strongly $\rho$-reachable} if  for any realization $\phi$, it holds that $ \prod_{i\in S_{\pi,\phi}} \delta_i \geq \rho$. That is, a strongly $\rho$-reachable  policy only adopts strongly $\rho$-reachable sequence. Let $\Omega(\rho^+)$ denote the set of all strong $\rho$-reachable policies. \end{definition}

We next introduce the term of maximal $\rho$-reachable sequence.

\begin{definition}[Maximal $\rho$-reachable Sequence]
\label{def:2}
Fix any $\rho\in[0,1]$.  When $\prod_{i\in E} \delta_i <  \rho$, we say a sequence  $S$ is a maximal $\rho$-reachable sequence if $S$ is $\rho$-reachable but not strongly  $\rho$-reachable. When  $\prod_{i\in E} \delta_i \geq  \rho$, all sequences are considered as maximal $\rho$-reachable sequences. Let $\mathcal{G}(\rho)$ denote the set of all maximal $\rho$-reachable sequences.  \end{definition}

Intuitively, when  $\prod_{i\in E} \delta_i <  \rho$, a maximal $\rho$-reachable sequence $G\in \mathcal{G}(\rho)$ must satisfy two conditions: 1. the reachability of all items in $G$ is at least $\rho$, and 2. any item placed after $G$ has reachability less than $\rho$.

Based on the above notations, we first show that there exists a $\rho$-reachable policy whose performance is close to the optimal policy.

\begin{lemma}
\label{lem:1}Fix any $\rho\in[0,1]$. If $f$ is adaptive cascade submodular, there is a $\rho$-reachable policy $\pi \in \Omega(\rho)$ such that
$f_{avg}(\pi) \geq (1-\rho) f_{avg}(\pi^{opt})$.
\end{lemma}
\emph{Proof:} The basic idea of our proof is to show that given  an optimal policy $\pi^{opt}$, we can discard those items whose reachability is small at the cost of a bounded loss. A similar result was proved in \citep{kempe2008cascade} for maximizing a linear utility function under the non-adaptive setting.

For any maximal $\rho$-reachable sequence $G\in \mathcal{G}(\rho)$,   let $\Pr[(G, \psi)]$  denote the probability that $G$ is a prefix of some sequence adopted by $\pi^{opt}$ while the states of $G$ is $\psi$, thus $\mathrm{dom}(\psi)=G$. Based on this notation, we can represent the expected utility of $\pi^{opt}$ as follows:
 \begin{eqnarray}
f_{avg}(\pi^{opt})&= \sum_{G\in \mathcal{G}(\rho)}\sum_{\Psi: \mathrm{dom}(\Psi)=G}\Pr[(G, \Psi)] g(G, \Psi) 
\end{eqnarray}
where
 \begin{eqnarray*}
 &&g(G, \Psi) = \sum_{k\in [|G|]} (1-\delta_{G^k})(\prod_{i\in G^{(k-1)}}\delta_i) f(G^{(k)}, \Psi) \\
  &&+ (\prod_{i\in G} \delta_i) (\max_{\pi\in\Omega(E\setminus G)} f_{avg}(\pi\cup \mathrm{dom}(\Psi)\mid \Psi)-h(\Psi))
\end{eqnarray*} denotes the expected utility of $\pi^{opt}$ conditioned on  $G$ is a prefix of some sequence adopted by $\pi^{opt}$ while the states of $G$  is $\Psi$. According to Definition \ref{def:1}, we have $\max_{\pi\in\Omega(E\setminus G)} f_{avg}(\pi\cup \mathrm{dom}(\Psi)\mid \Psi)-h(\Psi) \leq \max_{\pi\in\Omega(E\setminus G)} f_{avg}(\pi \mid \emptyset)-f(\emptyset, \emptyset)$ due to $\emptyset \subseteq \Psi$. Moreover,   $\max_{\pi\in\Omega(E\setminus G)} f_{avg}(\pi \mid \emptyset)-f(\emptyset, \emptyset)=\max_{\pi\in\Omega(E\setminus G)} f_{avg}(\pi)\leq \max_{\pi\in\Omega(E)} f_{avg}(\pi)= f_{avg}(\pi^{opt})$ where the inequality is due to $\Omega(E\setminus G) \subseteq \Omega( E)$. Then we have \begin{equation}
\label{eq:pp}
\max_{\pi\in\Omega(E\setminus G)} f_{avg}(\pi\cup \mathrm{dom}(\Psi)\mid \Psi)-h(\Psi) \leq f_{avg}(\pi^{opt})
\end{equation}
 It follows that
\begin{eqnarray*}
 &&g(G, \Psi) \leq \sum_{k\in [|G|]} (1-\delta_{G^k})(\prod_{i\in G^{(k-1)}}\delta_i) f(G^{(k)}, \Psi) + \\&&\quad\quad\quad\quad\quad\quad\quad\quad\quad\quad\quad\quad\quad\quad(\prod_{i\in G} \delta_i) f_{avg}(\pi^{opt})\\
 &&\leq \sum_{k\in [|G|]} (1-\delta_{G^k})(\prod_{i\in G^{(k-1)}}\delta_i) f(G^{(k)}, \Psi) + \rho f_{avg}(\pi^{opt})
\end{eqnarray*}
where the first inequality is due to (\ref{eq:pp}) and the second inequality is due to $G$ is a maximal $\rho$-reachable sequence.
Then we have
{\small \begin{eqnarray}
&&f_{avg}(\pi^{opt})= \sum_{G\in \mathcal{G}(\rho)}\sum_{\Psi: \mathrm{dom}(\Psi)=G}\Pr[(G, \Psi)] g(G, \Psi) \nonumber\\
&&\leq \sum_{G\in \mathcal{G}(\rho)}\sum_{\Psi: \mathrm{dom}(\Psi)=G} \Pr[(G, \Psi)] \nonumber\\
&& \times (\sum_{k\in [|G|]} (1-\delta_{G^k})(\prod_{i\in G^{(k-1)}}\delta_i) f(G^{(k)}, \Psi)) + \rho f_{avg}(\pi^{opt})\label{eq:1}
\end{eqnarray}}

%
%
%

Based on the above results, we next construct a $\rho$-reachable policy whose expected utility is at least $(1-\rho) f_{avg}(\pi^{opt})$. Assume the optimal policy $\pi^{opt}$ is given, we run $\pi^{opt}$ until  the last item whose reachability is smaller than $\rho$ is selected or the current selecting process is dead, whichever comes first. It is clear  that the above policy is $\rho$-reachable, and its expected utility is {\small\begin{displaymath}\displaystyle \sum_{G\in \mathcal{G}(\rho)}\sum_{\Psi: \mathrm{dom}(\Psi)=G}\Pr[(G, \Psi)](\sum_{k\in [|G|]} (1-\delta_{G^k})(\prod_{i\in G^{(k-1)}}\delta_i) f(G^{(k)}, \Psi))\end{displaymath}}
whose value is lower bounded by $(1-\rho) f_{avg}(\pi^{opt})$ according to (\ref{eq:1}). This finishes the proof of this lemma. $\Box$

According to Lemma \ref{lem:1}, there exists a $\rho$-reachable policy whose expected utility is at least $(1-\rho) f_{avg}(\pi^{opt})$, then the optimal $\rho$-reachable policy has the same performance bound. In particular, let $\pi^{opt1}$ denote the optimal solution to \textbf{P1}, we have the following lemma.

\begin{lemma}
\label{lem:5}
Fix any $\rho\in[0,1]$. If $f$ is adaptive cascade submodular, we have $f_{avg}(\pi^{opt1}) \geq (1-\rho) f_{avg}(\pi^{opt})$.
\end{lemma}


Lemma \ref{lem:5} allows us to put our focus on solving  \textbf{P1}.

\begin{center}
\framebox[0.25\textwidth][c]{
\enspace
\begin{minipage}[t]{0.25\textwidth}
\small
\textbf{P1:}
\emph{Maximize $f_{avg}(\pi)$}\\
\textbf{subject to:} $\pi\in  \Omega(\rho)$
\end{minipage}
}
\end{center}
\vspace{0.1in}

We next introduce a new optimization problem \textbf{P2} subject to only adopting strongly $\rho$-reachable sequences. Let $\pi^{opt2}$ denote the optimal solution to \textbf{P2}.

\begin{center}
\framebox[0.25\textwidth][c]{
\enspace
\begin{minipage}[t]{0.25\textwidth}
\small
\textbf{P2:}
\emph{Maximize $f_{avg}(\pi)$}\\
\textbf{subject to:} $\pi\in  \Omega(\rho^+)$
\end{minipage}
}
\end{center}
\vspace{0.1in}

  Notice that every strongly $\rho$-reachable sequence is also a  $\rho$-reachable sequence, similarly, every  strongly $\rho$-reachable policy is also a  $\rho$-reachable policy. Therefore, $f_{avg}(\pi^{opt2})$ is upper bounded by  $f_{avg}(\pi^{opt1})$.  However, we next show that  the gap between   $f_{avg}(\pi^{opt1})$ and $f_{avg}(\pi^{opt2})$ can be bounded.  For notational simplicity, we use $f(i, \Phi)$ to denote $f(\{i\}, \Phi)$.
 \begin{lemma}
 \label{lem:3}
 Denote by $i^*=\max_{i\in E} \mathbb{E}_{\Phi\sim p(\phi)}[f(i, \Phi)]$ a single item with the maximum expected utility. If $f$ is adaptive cascade submodular, then $f_{avg}(\pi^{opt2})+ \mathbb{E}_{\Phi\sim p(\phi)}[f(i^*, \Phi)]\geq f_{avg}(\pi^{opt1})$.
\end{lemma}
 \emph{Proof:} Assume  the optimal $\rho$-reachable policy $\pi^{opt1}$ is given, we next construct a strongly $\rho$-reachable policy $\pi'$ as follows:  we run $\pi^{opt1}$ until  it violates the strongly $\rho$-reachable constraint (the item whose addition to the current solution violates the constraint is not selected)  or the current selecting process is dead, whichever comes first. 
Observe that any sequence $S$ of length $k$ can be represented as $S^{(k-1)}\oplus S^k$, where $\oplus$ is the concatenation operator. Assume $S$ is a $\rho$-reachable sequence, we have $\prod_{i\in S^{(k-1)}} \delta_i \geq \rho$, thus $S^{(k-1)}$ is a strongly $\rho$-reachable sequence according to Definition \ref{def:2sequence}. Based on this observation, it is easy to verify that  given any full realization, $\pi^{opt1}$ selects at most one more item (the item that violates the strongly $\rho$-reachable constraint) than  $\pi'$. Due to the adaptive submodularity of $f$ (Lemma \ref{lem:4}), the expected marginal utility of that item is upper bounded by $\mathbb{E}_{\Phi\sim p(\phi)}[f(i^*, \Phi)]$.  It follows that $f_{avg}(\pi')+\mathbb{E}_{\Phi\sim p(\phi)}[f(i^*, \Phi)] \geq f_{avg}(\pi^{opt1})$. Because $\pi^{opt2}$ is the optimal strongly $\rho$-reachable policy, we have $f_{avg}(\pi^{opt2})+\mathbb{E}_{\Phi\sim p(\phi)}[f(i^*, \Phi)]\geq f_{avg}(\pi')+ \mathbb{E}_{\Phi\sim p(\phi)}[f(i^*, \Phi)] \geq f_{avg}(\pi^{opt1})$. This finishes the proof of this lemma.  $\Box$ 

According to Definition \ref{def:2policy}, a policy $\pi$ is strongly $\rho$-reachable   if $\forall \phi:  \prod_{i\in S_{\pi,\phi}} \delta_i \geq \rho$, which is equivalent to  $\forall \phi:  \sum_{i\in S_{\pi,\phi}} -\log \delta_i \leq -\log \rho$. Replacing $\pi\in  \Omega(\rho^+)$ by $\forall \phi:  \sum_{i\in S_{\pi,\phi}} -\log \delta_i \leq -\log \rho$,  we obtain an alternative formulation \textbf{P2.1} of \textbf{P2}.

\begin{center}
\framebox[0.45\textwidth][c]{
\enspace
\begin{minipage}[t]{0.45\textwidth}
\small
\textbf{P2.1:}
\emph{Maximize $f_{avg}(\pi)$}\\
\textbf{subject to:} $\forall \phi:  \sum_{i\in S_{\pi,\phi}} -\log \delta_i \leq -\log \rho$
\end{minipage}
}
\end{center}
\vspace{0.1in}

  To facilitate  the analysis of our proposed algorithm, we introduce another optimization problem \textbf{P3} by replacing the objective function $f_{avg}(\pi)$ in  \textbf{P2.1} using $\mathbb{E}_{\Phi\sim  p(\phi)}[f(S_{\pi, \Phi}, \Phi)]$.

\begin{center}
\framebox[0.45\textwidth][c]{
\enspace
\begin{minipage}[t]{0.45\textwidth}
\small
\textbf{P3:}
\emph{Maximize $\mathbb{E}_{\Phi\sim  p(\phi)}[f(S_{\pi, \Phi}, \Phi)]$}\\
\textbf{subject to:} $\forall \phi:  \sum_{i\in S_{\pi,\phi}} -\log \delta_i \leq -\log \rho$
\end{minipage}
}
\end{center}
\vspace{0.1in}

  Note that $\mathbb{E}_{\Phi\sim  p(\phi)}[f(S_{\pi, \Phi}, \Phi)]$ is the expected utility of a policy $\pi$ assuming that the selecting process is never dead.  
  Therefore, if  $f$ is adaptive monotone, then  $f_{avg}(\pi)\leq \mathbb{E}_{\Phi\sim p(\phi)}[f(S_{\pi, \Phi}, \Phi)]$ for any $\pi$. For notation simplicity, define $\overline{f}_{avg}(\pi) = \mathbb{E}_{\Phi\sim  p(\phi)}[f(S_{\pi, \Phi}, \Phi)]$.
 Then we have the following lemma.
\begin{lemma}
\label{lem:2}
Let $\pi^{opt3}$ denote the optimal solution to  \textbf{P3}. If $f$ is adaptive monotone, then $\overline{f}_{avg}(\pi^{opt3}) \geq f_{avg}(\pi^{opt2})$.
\end{lemma}

\begin{algorithm}[hptb]
\caption{Candidate Policy $\pi_B$}
\label{alg:LPP2}
\begin{algorithmic}[1]
\STATE $S=\emptyset; \psi=\emptyset$
\WHILE {the selecting process is live}
\STATE let $i_r=\arg\max_{i \in E}\Delta(i\mid \psi)/c(i)$;
\STATE add $i_r$ to $S$; $\psi=\psi\cup\{\phi_{i_r}\};$
\ENDWHILE
\RETURN $S$
\end{algorithmic}
\end{algorithm}

\begin{algorithm}[hptb]
\caption{Restricted Candidate Policy $\pi^{\mathrm{restricted}}_B$}
\label{alg:LPP1}
\begin{algorithmic}[1]
\STATE $S=\emptyset; B=-\log \rho; \psi=\emptyset$
\WHILE {the selecting process is live}
\STATE let $i_r=\arg\max_{i \in E}\Delta(i\mid \psi)/c(i)$;
\IF {$B-c(i_r)\geq 0$}
\STATE add $i_r$ to $S$;
\STATE  $\psi=\psi\cup\{\phi_{i_r}\}; B\leftarrow B-c(i_r)$;
\ELSE
\STATE add $i_r$ to $S$; \label{line:!}
\STATE break;
\ENDIF
\ENDWHILE
\RETURN $S$
\end{algorithmic}
\end{algorithm}

\subsection{Algorithm Design}
Now we are ready to present our adaptive greedy plus policy $\pi^{\mathrm{greedy}+}$.  Define $c(i)=-\log \delta_i$ as the \emph{virtual} cost of an item $i\in E$. Intuitively, an item with a higher continuation probability has a smaller virtual cost.  $\pi^{\mathrm{greedy}+}$ randomly picks one solution from the two candidates $\pi_A$ and $\pi_B$ such that $\pi_A$ is selected with probability $1-\frac{1}{\rho(1-1/e)+1}$ and $\pi_B$ is selected with probability  $\frac{1}{\rho(1-1/e)+1}$, where the value of $\rho$ will be decided later. We next describe $\pi_A$ and $\pi_B$ in details.

\textbf{Design of $\pi_A$.} Selecting a singleton  $i^*$ with the maximum expected utility as the first item, then follow an arbitrary sequence of the remaining items to select the next item until the selecting process is dead. Clearly, $f_{avg}(\pi_A) \geq \mathbb{E}_{\Phi\sim p(\phi)}[f(i^*, \Phi)]$ due to $f$ is adaptive monotone.

\textbf{Design of $\pi_B$.} Selecting items in a greedy manner as follows:
In each round $r$ of a live selecting process, $\pi_B$ selects an item $i_r$ with the largest ``benefit-to-cost'' ratio \[i_r=\arg\max_{i\in E}\Delta(i\mid \psi)/c(i)\] where $\Delta(i\mid \psi)=\mathbb{E}_{\Phi\sim p(\phi\mid \psi)}[f(i \cup \mathrm{dom}(\psi), \Phi)]-h(\psi)$ denotes the expected marginal benefit of $i$ conditioned on the current partial realization $\psi$. This process iterates until  the current selecting process is dead. A detailed description of $\pi_B$ is listed in Algorithm \ref{alg:LPP2}.


\subsection{Performance Analysis}
For the purpose of analysis, we first introduce a restricted version $\pi^{\mathrm{restricted}}_B$ (Algorithm \ref{alg:LPP1})  of the second candidate solution $\pi_B$  by enforcing a budget constraint $ -\log \rho$ on the total virtual cost of selected items. That is, in each round of a live selecting process,  $\pi^{\mathrm{restricted}}_B$ selects an item  that maximizes the ratio of the expected marginal benefit to the virtual cost, and this process continues until the total virtual cost of selected items is larger than $ -\log \rho$ or the current selecting process is dead. Note that $\pi^{\mathrm{restricted}}_B$  is very similar to the adaptive greedy algorithm proposed in \citep{golovin2011adaptive} except that $\pi^{\mathrm{restricted}}_B$ is allowed to violate the budget constraint by adding one more item. That is, the first item that violates the budget constraint is also selected by $\pi^{\mathrm{restricted}}_B$. As $\pi_B$ always selects items no less than $\pi^{\mathrm{restricted}}_B$,  $\pi_B$ can not perform worse than $\pi^{\mathrm{restricted}}_B$ due to the utility function $f$ is adaptive monotone. To analyze the performance of $\pi_B$, it suffice to give a lower bound on $f_{avg}(\pi^{\mathrm{restricted}}_B)$.  


Now we are ready to present the main theorem of this paper.

\begin{theorem}
Fix any $\rho\in[0,1]$. If $f$ is adaptive cascade submodular and adaptive monotone, then \[f_{avg}(\pi^{\mathrm{greedy}+}) \geq \frac{\rho(1-1/e)(1-\rho)}{\rho(1-1/e)+1}f_{avg}(\pi^{opt})\]
\end{theorem}

\emph{Proof:} We first build a relation between $f_{avg}(\pi^{\mathrm{restricted}}_B)$ and $\overline{f}_{avg}(\pi^{opt3})$. Because we assume $f$ is  adaptive monotone and adaptive cascade submodular, we have $f$ is adaptive submodular according to Lemma \ref{lem:4}. Therefore, \textbf{P3} is an adaptive submodular maximization problem subject to a budget constraint, where the cost of each item $i\in E$ is $c(i)$ and the budget constraint is $ -\log \rho$. 
According to \citep{tang2020influence,golovin2011adaptive}, the ``benefit-to-cost'' ratio based greedy algorithm achieves approximation ratio $1-1/e^{\frac{l}{B}}$ when maximizing an adaptive submodular and adaptive monotone function, where $l$ is the (expected) actual amount of budget consumed by the algorithm and $B$ is the budget constraint. This ratio is lower bounded by $1-1/e$ when $l \geq B$. In our case, because $\pi^{\mathrm{restricted}}_B$ is allowed to violate the budget constraint by adding one more item to the solution, we have
\begin{equation}
\label{eq"2}
\overline{f}_{avg}(\pi^{\mathrm{restricted}}_B) \geq (1-1/e)\overline{f}_{avg}(\pi^{opt3})
 \end{equation}
Moreover, because $\pi^{\mathrm{restricted}}_B$ does not violate the budget constraint until the last round, the reachability of every item selected by $\pi^{\mathrm{restricted}}_B$  is lower bounded by $\rho$. Thus the expected utility  of $\pi^{\mathrm{restricted}}_B$ is at least $f_{avg}(\pi^{\mathrm{restricted}}_B)\geq \rho \overline{f}_{avg}(\pi^{\mathrm{restricted}}_B)$. It follows that
\begin{equation}
f_{avg}(\pi^{\mathrm{restricted}}_B)\geq \rho(1-1/e)\overline{f}_{avg}(\pi^{opt3}) \label{eq:2}
 \end{equation}due to (\ref{eq"2}). Now we are ready to bound the approximation ratio of $\pi^{\mathrm{greedy+}}$. Let $\alpha(\rho)=\frac{1}{\rho(1-1/e)+1}$ and $\overline{\alpha}(\rho)=1-\alpha(\rho)$, we have
\begin{eqnarray}&&f_{avg}(\pi^{\mathrm{greedy+}})
=\overline{\alpha}(\rho)f_{avg}(\pi_A)+\alpha(\rho)f_{avg}(\pi_B) \label{Eq:11}\\
&\geq&\alpha(\rho)f_{avg}(\pi^{\mathrm{restricted}}_B)+\overline{\alpha}(\rho)\mathbb{E}_{\Phi\sim p(\phi)}[f(i^*, \Phi)] \label{Eq:1xxx}\\
&\geq& \rho(1-\frac{1}{e})\alpha(\rho)\overline{f}_{avg}(\pi^{opt3})+\overline{\alpha}(\rho)\mathbb{E}_{\Phi\sim p(\phi)}[f(i^*, \Phi)] \label{Eq:1}\\
&\geq&  \rho(1-\frac{1}{e})\alpha(\rho)(f_{avg}(\pi^{opt2})+\mathbb{E}_{\Phi\sim p(\phi)}[f(i^*, \Phi)]) \label{Eq:2}\\
&\geq&  \rho(1-\frac{1}{e})\alpha(\rho) f_{avg}(\pi^{opt1}) \label{Eq:4}\\
&\geq& (1-\rho)\rho(1-\frac{1}{e})\alpha(\rho)f_{avg}(\pi^{opt}) \label{Eq:5}
\end{eqnarray}
(\ref{Eq:11}) is due to $\pi^{\mathrm{greedy}^+}$ randomly picks one between $\pi_A$ and $\pi_B$ such that $\pi_A$ is selected with probability $1-\alpha(\rho)$ and $\pi_B$ is selected with probability  $\alpha(\rho)$; (\ref{Eq:1xxx}) is due to $f_{avg}(\pi^{\mathrm{restricted}}_B)\leq {f}_{avg}(\pi_B)$ and ${f}_{avg}(\pi_A) \geq \mathbb{E}_{\Phi\sim p(\phi)}[f(i^*, \Phi)]$; (\ref{Eq:1}) is due to (\ref{eq:2}); (\ref{Eq:2}) is due to Lemma \ref{lem:2}; (\ref{Eq:4}) is due to Lemma \ref{lem:3}; (\ref{Eq:5}) is due to Lemma \ref{lem:5}. This finishes the proof of this theorem. $\Box$

It can be seen that if we set $\rho=\frac{\sqrt{e(2e-1)}-e}{e-1}$, then $(1-\rho)\rho(1-1/e)\alpha(\rho)>0.12$. Hence,  $\pi^{\mathrm{greedy+}}$ achieves  a $0.12$ approximation ratio at  $\rho=\frac{\sqrt{e(2e-1)}-e}{e-1}$ .

\begin{corollary}
\label{thm:adap1}
If we set $\rho=\frac{\sqrt{e(2e-1)}-e}{e-1}$, then our adaptive greedy plus policy $\pi^{\mathrm{greedy+}}$ achieves a $0.12$ approximation ratio  given that $f$ is adaptive cascade submodular and adaptive  monotone.
\end{corollary}

\section{Performance Evaluation}
\label{sec:experiment}
\begin{figure*}[hptb]
\hspace*{-0.75in}
\includegraphics[scale=0.21]{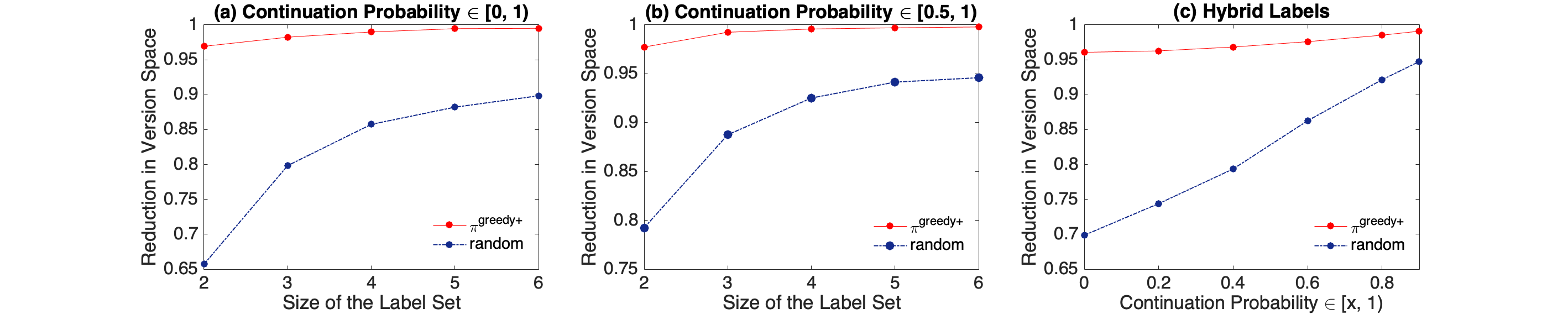}
\caption{Reduction in version space vs. number of possible labels for each query.}
\label{fig:puri_reduction}
\end{figure*}

\begin{figure*}[hptb]
\hspace*{-0.75in}
\includegraphics[scale=0.21]{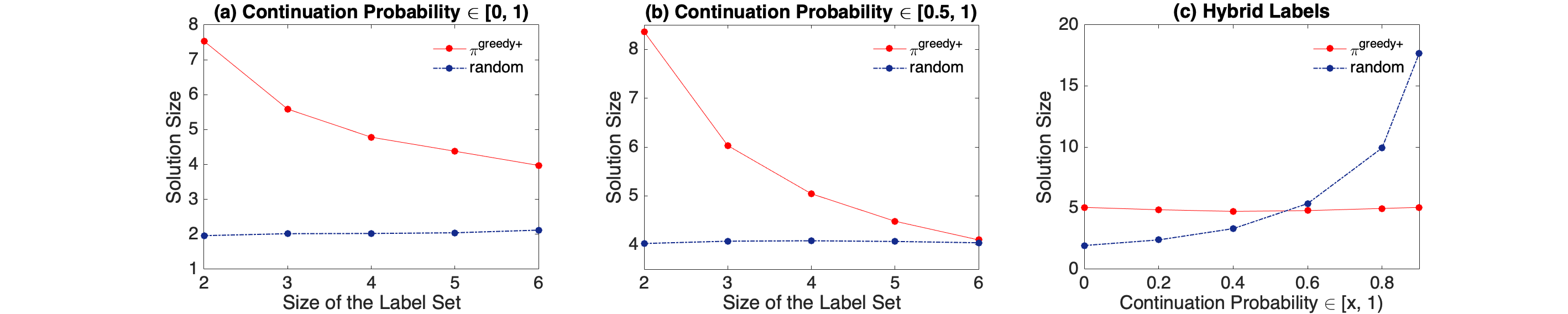}
\caption{(a) Reduction in version space vs. lower end of the sample range of the continuation probability for each query; and (b) Statistics of the percentage of the selected queries in each group.}
\label{fig:hybrid_reduction_count}
\end{figure*}

\begin{figure*}[hptb]
\hspace*{-0.75in}
\includegraphics[scale=0.21]{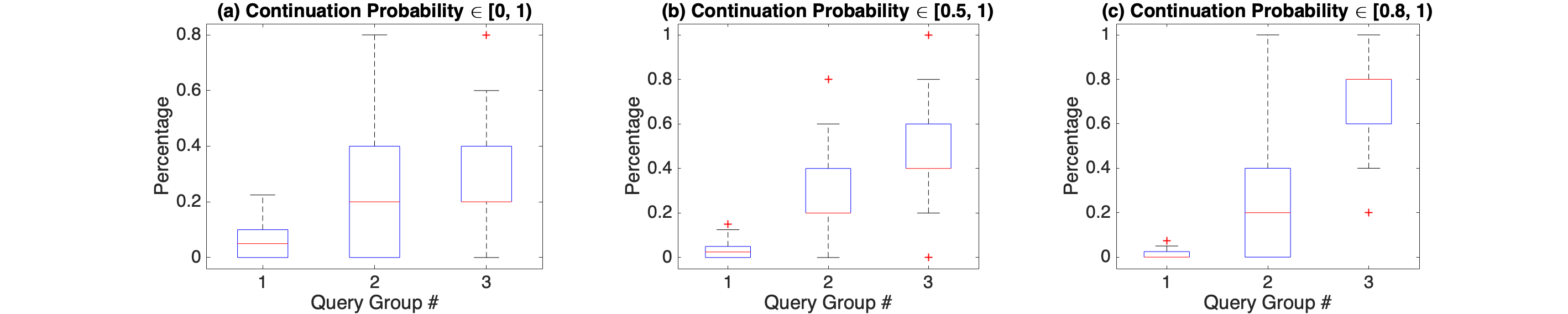}
\caption{Statistics of the percentage of the selected queries in each group.}
\label{fig:hybrid_percentage}
\end{figure*}
In this section, we conduct experiments to evaluate the performance of our proposed algorithm in the context of active learning. Assume that we are given a set of hypotheses $H$, and a set of unlabeled data points $X$ where each $x\in X$ is drawn independently from some distribution $D$. In pool-based active learning, in order to avoid the cost of obtaining labeled data from domain experts, we adaptively identify a sequence of queries that labels a few unlabeled examples until the labels of all unlabeled examples are implied by the obtained labels. Each query is associated with a continuation probability. In the context of online survey (the third example introduced in Section \ref{sec:intro}), the continuation probability of a query (a survey question) represents the likelihood of a user continuing to answer the next query after answering the current one.

We define the version space to be the set of hypotheses consistent with the observed labels. Intuitively our goal is to reduce the probability mass of the version space as much as possible. Reducing the version space amounts to filtering false hypotheses with stochastic queries. Note query $x\in X$ filters all hypotheses that disagree with the target hypotheses $h^*$ at $x$. The reduction in version space mass is shown to be adaptive submodular \citep{golovin2011adaptive}.
When running $\pi^{\mathrm{greedy}+}$, we measure the conditional marginal utility of a query as the expected reduction in version space based on the observation of previous requested labels.

In our experiments, we consider $1000$ hypotheses with $50$ unlabeled data points. For each hypothesis, its probability is drawn from $(0,1)$ uniformly at random with unity generalization; each data point is assigned a value randomly selected from its possible set of labels. 
 All experiments were run on a machine with Intel Xeon 2.40GHz CPU and 64GB memory, running 64-bit RedHat Linux server. For each set of experiments, we run the simulation for $1,000$ rounds and report the average results as follow.

Our first set of experiments evaluate the performance of our algorithm as measured by the yielded reduction in version space with respect to the changes in the number of possible labels for each data point. We use a random algorithm as our baseline. The random algorithm outputs a sequence of queries in an adaptive manner where each query is randomly selected until the current selecting process is dead. We consider the scenario where each data point has the same number of possible labels. We vary the size of the label set and measure how the performance of our algorithm changes under different settings.

The results are plotted in Figure \ref{fig:puri_reduction}(a) and \ref{fig:puri_reduction}(b). As shown in the figures, the $x$-axis refers to the size of the label set, ranging from two to six. The $y$-axis refers to the reduction in version space generated by the corresponding algorithms. We consider two settings of the continuation probability for each query. Figure \ref{fig:puri_reduction}(a) shows the results where each query is assigned a continuation probability drawn uniformly at random from $[0, 1)$. We observe that as the size of the label set increases, the reduction in version space increases for both algorithms. We also observe that $\pi^{\mathrm{greedy}+}$ yields $96.931\%$ reduction in version space for binary labels. It achieves $99.505\%$ reduction when data points have $6$ possible labels. It also significantly outperforms the baseline random algorithm in all test cases. The random algorithm only yields $65\%$ reduction in the case of binary labels. Note that our algorithm considers the marginal utility as well as the continuation probability for each query, leading to a higher reduction in version space. In Figure \ref{fig:puri_reduction}(b), each query is assigned a continuation probability drawn uniformly at random from $[0.5, 1)$. Similarly, we observe that $\pi^{\mathrm{greedy}+}$ yields $97.729\%$ reduction in the case of binary labels, and achieves $99.786\%$ reduction when data points have $6$ possible labels. Again it outperforms the baseline significantly.

Our second set of experiments explore the impact of the continuation probability of the queries on the reduction in version space, as illustrated in Figure \ref{fig:puri_reduction}(c). We consider the scenario where data points have various numbers of possible labels. We randomly divide our $50$ unlabeled data points into three groups. The first group contains $40$ data points with binary labels. The second group contains $5$ data points with three possible labels. The third group contains $5$ data points with four possible labels. We vary the lower end of the interval from which we sample the continuation probability of queries, ranging from $0$ to $0.9$. Specifically, if the lower end is set to $0$, we sample the continuation probability uniformly at random from $[0,1)$; if the lower end is set to $0.6$, we sample the continuation probability uniformly at random from $[0.6, 1)$.

As shown in Figure \ref{fig:puri_reduction}(c), the $x$-axis holds the lower end of the sample interval, and the $y$-axis holds the reduction in version space generated by the corresponding algorithms. We observe that as expected, the reduction in version space increases as the lower end of the sample interval increases. Intuitively a larger lower end indicates that the continuation probability is sampled from a pool of larger values. Therefore more queries can be selected in the output sequence, leading to a higher reduction in version space.

Our next set of experiments evaluate how the solution size changes with respect to the changes in the size of the label set, under the scenario where each data point has the same number of possible labels. We vary the size of the label set and measure how the solution size changes under different settings. Figure \ref{fig:hybrid_reduction_count}(a) and \ref{fig:hybrid_reduction_count}(b) show the  results where each query is assigned a continuation probability drawn uniformly at random from $[0, 1)$ and $[0.5, 1)$, respectively. We observe that as the size of the label set increases, the size of the solution generated by $\pi^{\mathrm{greedy}+}$ decreases. We also observe that the size of the solution generated by the random algorithm does not change with the size of the label set.

Next we explore the impact of the continuation probability of the queries on the solution size, as illustrated in Figure \ref{fig:hybrid_reduction_count}(c). We use the same setting as in Figure \ref{fig:puri_reduction}(c). We observe that as the sample range of the continuation probability narrows down to the large values, the size of the solution generated by the random algorithm increases rapidly. For $\pi^{\mathrm{greedy}+}$, the solution size does not change much with the sample range.

We take a closer look at the composition of the queries in the output sequence generated by $\pi^{\mathrm{greedy}+}$. As aforementioned we have three groups of queries. We measure the percentage of selected queries in each group. For example, out of $40$ data points with binary labels, if $2$  of them are selected, then $2/40=5\%$ of the queries in group $1$ are selected. We plot the statistics for each query group in Figure \ref{fig:hybrid_percentage}. The $x$-axis refers to the group number of the queries, and the $y$-axis refers to the percentage of the selected queries in the corresponding group. Figure \ref{fig:hybrid_percentage}(a), \ref{fig:hybrid_percentage}(b) and \ref{fig:hybrid_percentage}(c) plot the results as the continuation probability of each query is sampled from $[0, 1)$, $[0.5, 1)$, $[0.8, 1)$, respectively. We observe that as the sample range narrows down to the larger values, more queries in group $3$, and less queries in group $1$, are selected. The reason is that $\pi^{\mathrm{greedy}+}$ takes into account both conditional marginal utility and the continuation probability for each candidate query. For the case with sample interval $[0, 1)$, while queries with four possible labels tend to yield a higher marginal utility, all of them may not be associated with a large continuation probability. Choosing one with small continuation probability may diminish the chance of choosing more queries to further reduce the version space. Therefore some queries with four possible labels are displaced in the output sequence in favor of queries with binary labels and large continuation probability. As the lower limit of the continuation probability goes up, more queries with four possible labels are included in the output sequence.

\section{Conclusion}
In this paper, we propose and study a new stochastic optimization problem, called adaptive cascade submodular maximization. Our goal is to adaptively select a sequence of items that maximizes the expected utility. Our problem is motivated by many real-world applications  where the selecting process could be terminated prematurely. We show that existing studies on submodular maximization do not apply to our setting. We start by introducing a class of stochastic utility functions, adaptive cascade submodular functions. Then we propose an adaptive policy that achieves a constant approximation ratio given that the utility function is adaptive cascade submodular and adaptive monotone. In the future, we would like to extend this work by incorporating some practical constraints such as cardinality constraint to the existing model.

\bibliographystyle{ijocv081}
\bibliography{reference}




\end{document}